\documentclass[]{spie}  

\usepackage{amsmath,amsfonts,amssymb}
\usepackage{graphicx}
\usepackage[colorlinks=true, allcolors=blue]{hyperref}
\usepackage{subfigure}
\usepackage{algorithmic}
\usepackage{algorithm}
\usepackage{amsmath}

\title{KLO-Net: A Dynamic K-NN Attention U-Net with CSP Encoder for Efficient Prostate Gland Segmentation from MRI\footnotemark[2]}

\author[1]{Anning Tian}
\author[1]{Byunghyun Ko}
\author[1]{Kaichen Qu}
\author[1]{Mengyuan Liu}
\author[1,*]{Jeongkyu Lee}
\affil[1]{Northeastern University, 4 N 2nd St, San Jose, USA}

\authorinfo{A.Tian, B.Ko, K.Qu, M.Liu, J.Lee are with the Khoury College of Computer Sciences, Northeastern University, San Jose, CA, 95113, USA. *Corresponding author: jeo.lee@northeastern.edu}

\begin{document} 
\maketitle

\footnotetext[2]{Preprint. Accepted to \textit{SPIE Medical Imaging 2026: Image Processing}.}

\begin{abstract}

Real-time deployment of prostate MRI segmentation on clinical workstations is often bottlenecked by computational load and memory footprint. Deep learning-based prostate gland segmentation approaches remain challenging due to anatomical variability. To bridge this efficiency gap while still maintaining reliable segmentation accuracy, we propose KLO-Net, a dynamic K-Nearest Neighbor attention U-Net\cite{Ronneberger15} with Cross Stage Partial\cite{Wang20}, i.e., CSP, encoder for efficient prostate gland segmentation from MRI scan. Unlike the regular K-NN attention mechanism\cite{wang21}, the proposed dynamic K-NN attention mechanism allows the model to adaptively determine the number of attention connections for each spatial location within a slice. In addition, CSP blocks address the computational load to reduce memory consumption. To evaluate the model's performance, comprehensive experiments and ablation studies are conducted on two public datasets, i.e., PROMISE12\cite{Litjens14} and PROSTATEx\cite{Litjens17}, to validate the proposed architecture. The detailed comparative analysis, i.e., KLO-Net against five state-of-the-art methods, demonstrates the model's advantage in computational efficiency and segmentation quality.

\end{abstract}

\section{INTRODUCTION}
\label{sec:intro}
Accurate segmentation of the prostate gland from magnetic resonance imaging (MRI) is extremely important in the diagnosis, treatment planning, and monitoring of prostate cancer. According to the American Cancer Society, an estimated 35,770 men in the United States are expected to die from prostate cancer, and prostate cancer continues to rank as one of the most prevalent cancers among men worldwide\cite{Siegel25}. Prostate segmentation presents unique challenges that other target structures do not, mainly due to its variable shape, ambiguous boundaries, and anatomical differences between patients\cite{Litjens14}. Segmentation of medical images using convolutional neural networks (CNNs) has become increasingly prevalent in recent years, and the U-Net\cite{Ronneberger15} remains the dominant architecture for medical image segmentation.

While the standard U-Net effectively captures both semantic and spatial information in most target structures, it still relies heavily on convolutional operations with limited receptive fields. This means that the standard U-Net may fail to capture long-range dependencies, which may be crucial for complex anatomies like the prostate gland. Attention mechanisms, particularly self-attention, have enabled U-Net-based architectures to capture global context\cite{Petit21}. However, the high computational costs, especially in medical images that tend to be high resolution, continue to be a great challenge for clinical deployment\cite{Liu2024VSmTrans}.

To address these challenges and limitations, sparse attention mechanisms, such as K-Nearest Neighbor (K-NN) attention modules, have been proposed\cite{wang21}. K-NN attention allows the model to retain the contextual modeling ability while reducing the quadratic complexity of self-attention modules. However, despite these advances, many existing implementations for prostate segmentation utilize a fixed number of neighbors.

Cross-Stage Partial (CSP) networks\cite{Wang20} have also gained popularity due to their ability to reduce computational redundancy through partial feature transformation. As medical image segmentation approaches require not only accuracy but also efficiency to be realistic clinical solutions, computational efficiency has become increasingly crucial. CSP networks have shown considerable promise in reducing computational cost. However, CSP modules remain an underexplored topic in the field of medical image segmentation.

In this paper, we propose KLO-Net, a novel U-Net-based architecture that utilizes a dynamic K-Nearest Neighbor (K-NN) attention mechanism with CSP encoder blocks to achieve efficient and accurate prostate gland segmentation from MRI. Unlike the regular K-NN attention mechanism, the proposed dynamic K-NN attention mechanism allows the model to adaptively determine the number of attention connections for each spatial location within a slice. In addition, CSP blocks address the computational load to reduce memory consumption.

To evaluate the model's performance, comprehensive experiments and ablation studies are conducted on two public datasets, i.e., PROMISE12\cite{Litjens14} and PROSTATEx\cite{Litjens17}, to validate the proposed architecture. We find that KLO-Net substantially reduces computational demands and model complexity while achieves higher segmentation accuracy.

\section{RELATED WORK}

\subsection{U-Net for Medical Image Segmentation}

Since the introduction of U-Net\cite{Ronneberger15}, it has become the cornerstone architecture for medical image segmentation. The U-Net architecture effectively captures both contextual and local information based on its U-shaped encoder-decoder structure with skip connections. Numerous extended works of U-Net have been proposed since then, including 3D U-Net\cite{Cicek16}, V-Net\cite{Milletari16}, and nnU-Net\cite{Isensee21}. Attention U-Net\cite{Oktay18} first introduces the attention mechanisms to the U-Net architecture. It incorporated attention gates to focus on salient regions, thereby enhancing the performance of U-Net. Cross-Slice Attention Module (CSAM) \cite{Hung24}, a 2.5D U-Net, extends attention to inter-slice volumetric data by capturing partial spatial information to improve segmentation performance. TransUNet\cite{Chen21} combines U-Net with Vision Transformer blocks, utilizing self-attention to model the global context while keeping the localization ability of convolutional features.

\subsection{U-Net Variants for Prostate Gland Segmentation}
Prostate gland segmentation has been a popular domain for breakthroughs and remains a crucial task in medical image analysis. Various U-Nets and their variants have been published in previous works due to the importance of prostate gland segmentation. While U-Net variants remain popular in prostate gland segmentation tasks due to their ability to fine-tune anatomical boundaries, transformer networks combined with U-Net's traditional encoder-decoder structures have also been used in prostate gland segmentation studies. 

Some notable prostate segmentation models include CSAM \cite{Hung24}, CAT-Net \cite{CAT_Hung23}, and the model proposed by Liu et al. \cite{Liu2020}. CAT-Net introduces a Cross-Slice Attention Transformer module, which allows the model to learn inter-slice dependencies across the entire prostate volume, thereby improving consistency in zonal segmentation. CSAM presents a 2.5D cross-slice attention module that aggregates information across all slices in anisotropic MRI volumes to enhance volumetric context modeling for segmentation. Liu et al. propose a Bayesian deep attentive neural network. The proposed architecture leverages spatial attention and uncertainty estimation, utilizing a multi-scale feature pyramid attention and a tuned ResNet-50 backbone. These methods demonstrate the growing trend towards using attention mechanisms and hybrid network designs to improve anatomical context modeling and segmentation accuracy.

\subsection{Sparse Attention Mechanisms}

Self-attention mechanisms have become the backbone of many state-of-the-art segmentation models\cite{Chen21,Hatamizadeh22,Cao21}; however, their quadratic computational complexity brings the challenge to the efficiency cost. Self-attention, proposed by Vaswani et al.\cite{Vaswani17}, computes interactions between all position pairs to enable global context modeling. When the mechanism is applied to medical images, the quadratic computational complexity increases the computation costs; a $512 \times 512$ image will require over 68 billion pairwise computations, making real-time clinical deployment infeasible.

To address the issue, sparse attention mechanisms have been proposed. K-Nearest Neighbor (K-NN) attention\cite{wang21} provides an adaptive approach. By computing attention only between each query and its k most relevant keys based on similarity scores, K-NN attention reduces computational complexity from $O(N^2)$ to $O(N k \log N)$ while maintaining the ability to capture important interactions. K-NN attention can approximate full self-attention with an appropriate k selection\cite{Bhojanapalli21}.

\subsection{Cross Stage Partial Networks}

Cross Stage Partial (CSP) networks\cite{Wang20} introduce a strategy to enhance gradient flow and reduce computational redundancy in deep neural networks. In a CNN, it splits feature maps into two pathways: (i) transformation and (ii) identity mapping, and then merging them at the end to reduce memory consumption. This design preserves gradient flow, reduces memory consumption, and computational cost. Since its proposal, CSP has been widely adopted in state-of-the-art object detection networks, including YOLOv4\cite{Bochkovskiy20} with CSPDarknet backbone and YOLOv5\cite{Jocher20}.

\section{METHODS}
\subsection{KLO-Net Model Architecture}

\begin{figure}[t]
    \centering
    \includegraphics[width=0.99\linewidth]{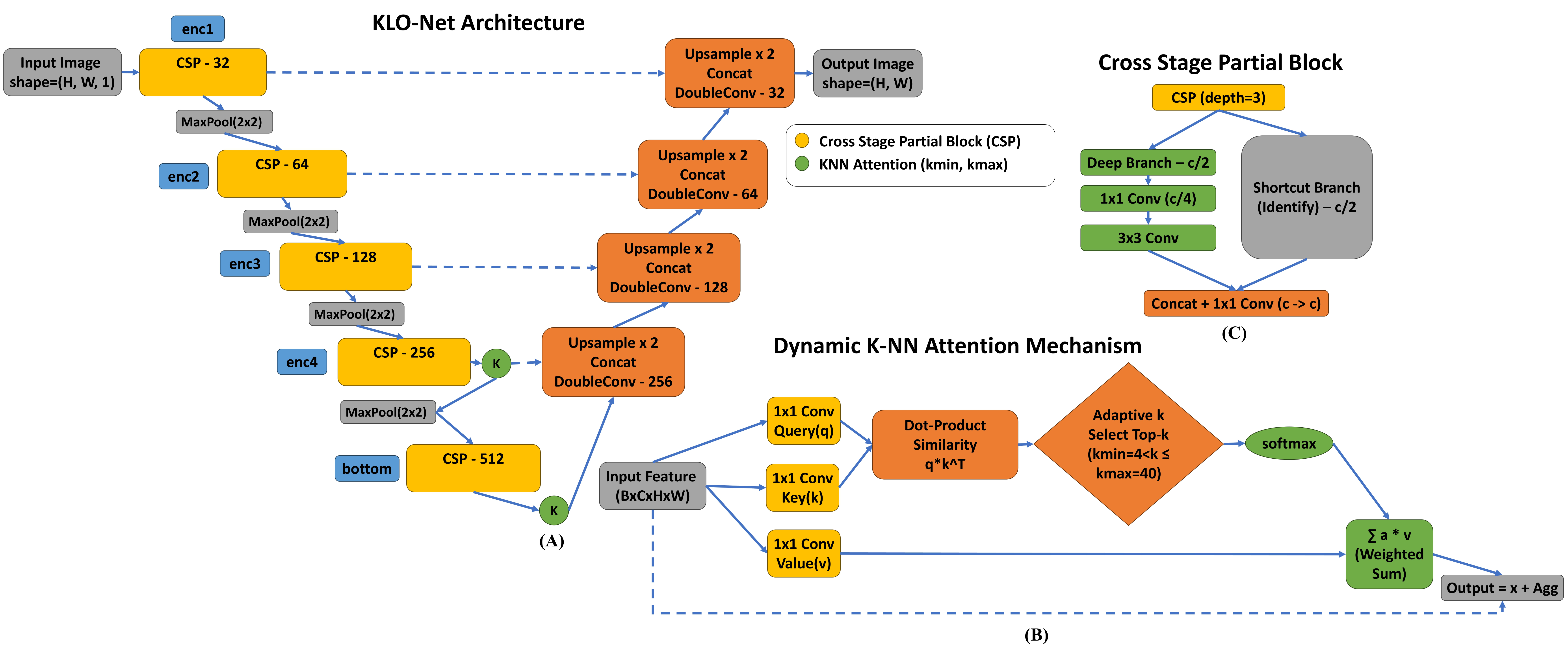}
    \caption{Architecture of KLO-Net: (A) KLO-Net architecture, (B) Dynamic K-NN Attention, and (C) CSP block design.}
    \vspace{3pt}
    \label{fig:architecture}
\end{figure}

KLO-Net, as shown in Fig.~\ref{fig:architecture}(A), is a U-Net style encoder–decoder network tailored for prostate gland segmentation on MRI. The encoder replaces the standard double-convolution blocks with Cross Stage Partial (CSP) modules to improve parameter efficiency, while the decoder retains the original U-Net design with skip connections and upsampling. On the deepest encoder stage and the bottleneck, a dynamic K-NN attention modules Fig.~\ref{fig:architecture}(B) is inserted to capture long-range context using sparse, content-adaptive attention. The CSP block structure used in KLO-Net is illustrated in Fig.~\ref{fig:architecture}(C).

\subsection{Dynamic K-Nearest Neighbor Attention}

Accurate prostate segmentation benefits from modeling interactions between distant regions, e.g., across opposite sides of the gland boundary. However, applying full self-attention on high-resolution feature maps is computationally expensive. Our Dynamic K-NN attention module addresses this problem by sparsifying the attention graph and adaptively controlling, for each spatial location, how many neighbors to attend to. The following subsections describe the dynamic $k$ selection mechanism and the resulting sparse attention formulation in detail.

\subsubsection{Dynamic K Selection Mechanism}

K-NN attention\cite{wang21} reduces the quadratic complexity of self-attention by computing attention weights only between each query and its k most similar keys. Given similarity scores
\begin{equation}
S \in \mathbb{R}^{B \times h \times N \times N}
\end{equation}
K-NN attention selects top-k neighbors for each query position, thereby restricting each query to at most k non-zero attention connections and substantially alleviating the quadratic cost of standard self-attention. However, this approach uses a fixed k value across all spatial positions within the slice and may be suboptimal for medical images with varying regional complexity.

Our proposed dynamic K-NN attention extends K-NN attention by introducing a Dynamic K selection mechanism. The detailed structure of dynamic K-NN attention mechanism is illustrated in Fig.~\ref{fig:architecture}(B). Dynamic K-NN attention mechanism predicts a position-specific parameter $\tau \in (0,1)$ where $\tau$ (tau) represents the attention density required at each spatial location. Values of $\tau$ close to 1 indicate regions requiring dense attention connections, such as organ boundaries, while values close to 0 suggest sparse connections are sufficient, typically for uniform background areas. The number of neighbors is computed as:
\begin{equation}
k = \max(k_{\min}, \min(k_{\max}, \lfloor\tau \cdot k_{\max}\rfloor))
\label{eq:number_of_neighbors}
\end{equation}
where $k_{\min}$ and $k_{\max}$ are manually set boundaries to ensure $k$ remains within the range while adapting to local image content.

\subsubsection{Feature Projection}

Query, key, and value representations are generated through learned linear transformations implemented as $1 \times 1$ convolutions. After the input feature map $x \in \mathbb{R}^{C \times H \times W}$ is received, we compute $Q = W_q(x)$, $K = W_k(x)$, and $V = W_v(x)$, where $W_q, W_k, W_v \in \mathbb{R}^{C \times C}$. For multi-head attention, these tensors are reshaped to $\mathbb{R}^{B \times h \times N \times d_h}$ format, where $B$ represents batch size, $h$ the number of heads, $N = H \times W$ the number of spatial positions, and $d_h = C/h$ the dimension per head.

\subsubsection{Similarity Computation}

The similarity score between pairwise queries and keys is computed using the standard scaled dot-product attention formulation\cite{Vaswani17} $S_{ij} = q_i^T k_j / \sqrt{d_h}$, where the scaling factor $\sqrt{d_h}$ prevents gradient vanishing. This yields the full similarity matrix $S \in \mathbb{R}^{B \times h \times N \times N}$, ensuring similarity scores remain within a stable range.

\subsubsection{Dynamic k Prediction}

Our novel gating network $G$ predicts $\tau$ for each spatial location:
\begin{equation}
\tau = \sigma(G(x)) \in \mathbb{R}^{B \times 1 \times H \times W}
\end{equation}
where $\sigma$ is the sigmoid function and the gating network consists of two $1 \times 1$ convolutions with a ReLU activation:
\begin{equation}
G(x) = \text{Conv}_2(\text{ReLU}(\text{Conv}_1(x)))
\end{equation}
where $\text{Conv}_1: \mathbb{R}^C \rightarrow \mathbb{R}^{C/4}$ performs dimensionality reduction with $1 \times 1$ convolution, and $\text{Conv}_2: \mathbb{R}^{C/4} \rightarrow \mathbb{R}^1$ projects to a single-channel complexity map. The intermediate ReLU activation enables the modeling of nonlinear complexity. The design adds minimal parameters $(C^2/4 + C/4)$ while effectively capturing spatial complexity variations.

\subsubsection{Sparse Attention Computation}

For each query position $i$, its $k_i$ nearest neighbors are selected based on similarity scores, where $k_i$ is computed using the previously defined dynamic selection mechanism, equation ~\eqref{eq:number_of_neighbors}. We identify the top-$k_i$ most similar keys through $N_i = \text{TopK}(S_{i,:}, k_i)$, where $N_i$ contains the indices of selected neighbors. The attention weights are then computed using masked softmax, where only the selected neighbors contribute to the final output. Specifically, softmax is applied to the similarity scores multiplied by a binary mask $M$, where $M_{ij} = 1$ if $j \in N_i$ and 0 otherwise. The final output aggregates values from selected neighbors through weighted summation: $o_i = \sum_j A_{ij} v_j$.

After reshaping back to spatial dimensions and applying a projection layer, the output feature map is obtained.

\subsection{Cross Stage Partial Block}

Cross Stage Partial (CSP)\cite{Wang20} is a network design strategy. By splitting feature maps into parallel paths, it is able to reduce computational redundancy. While CSP is being widely implemented in object detection, its potential for medical image segmentation remains unexplored. Specifically, the CSP module implemented by KLO-Net utilizes bottleneck structures, which is termed CSP modules. The architecture of CSP blocks used in KLO-Net is shown in Fig.~\ref{fig:architecture}(C).

In KLO-Net, the standard double convolution blocks in U-Net's encoder have been replaced by CSP modules. Given input features $x \in \mathbb{R}^{C \times H \times W}$, two parallel $1\times1$ convolutions project $x$ to $x_1, x_2 \in \mathbb{R}^{C_h \times H \times W}$, where $C_h$ is a reduced channel width (typically $C/2$). The first branch $x_1$ is processed by $n$ sequential bottleneck blocks (each: $1\times1$ channel reduction $\rightarrow 3\times3$ spatial conv), while the second branch $x_2$ serves as a lightweight shortcut only apply a $1\times1$ projection. The transformed $x_1'$ and $x_2$ are concatenated and fused by a final $1\times1$ convolution to produce the output. All DoubleConv blocks are replaced with CSP at channel depths $32 \rightarrow 64 \rightarrow 128 \rightarrow 256 \rightarrow 512$, preserving accuracy while improving efficiency.

\subsection{Architecture Design}

KLO-Net follows the U-Net encoder-decoder architecture, comprising five encoding stages, four decoding stages, and one bottleneck layer. CSP is utilized to replace the standard double convolution encoder blocks at each level, while the decoder retains the original U-Net design, featuring skip connections and upsampling. The key components of KLO-Net are as follows:

\textbf{Input:} Single-channel grayscale MRI slices of size $(H \times W \times 1)$ are fed into the network.

\textbf{Encoder:} Five encoding stages progressively downsample the spatial resolution while increasing the number of feature channels. Each encoder stage, except the initial level, consists of a MaxPool2d layer followed by a Bottleneck CSP block. The dynamic K-NN attention module is inserted after the CSP at encoder four and bottleneck.

\textbf{Bottleneck:} The deepest layer processes features at 1/16 of the original resolution, utilizing 512 channels that incorporate both CSP and Dynamic K-NN attention.

\textbf{Decoder:} Four upsampling stages symmetrically restore the spatial resolution. Each decoder stage performs bilinear upsampling (scale factor=2), concatenates with the corresponding encoder features via skip connections, and applies double convolution blocks (Conv-BN-ReLU $\times$ 2).

\textbf{Output:} A final $1 \times 1$ convolution projects the 32-channel features to num\_classes=2 for binary segmentation.

In our experiments, we apply Dynamic K-NN attention to encoder four and bottleneck with csp\_depth=3 for optimization.

\subsection{Loss Function}

For KLO-Net training, a combination of Dice loss and Boundary loss\cite{KERVADEC2021101851} is implemented, which can balance region-based and boundary-based optimization for prostate segmentation.

The Dice loss\cite{Milletari16} ensures overall segmentation accuracy by maximizing the overlap between the predicted and ground-truth regions. It is defined as $\mathcal{L}_{\text{Dice}} = 1 - 2|P \cap G|/(|P| + |G|)$, where $P$ and $G$ represent the predicted and ground truth regions respectively. The Boundary loss complements the Dice loss by explicitly optimizing the segmentation boundaries for precise prostate delineation. This loss computes the mean of pixel-wise products between the predicted probabilities $s_\theta$ and the signed distance map $\phi_G$ of the ground truth.

For ablation studies, the following loss function employed:
\begin{equation}
\mathcal{L}_{\text{ablation}} = 0.9 \cdot \mathcal{L}_{\text{Dice}} + 0.1 \cdot \mathcal{L}_{\text{Boundary}}
\end{equation}

This combination provides a unified baseline across all ablation experiments, ensuring that performance differences are attributed solely to architectural modifications, rather than variations in the loss function.

For comparison with state-of-the-art methods, we adopt the Focal Tversky Loss\cite{Abraham19} to optimize performance on highly imbalanced data:
\begin{equation}
\mathcal{L}_{\text{comparison}} = (1 - TI)^{\gamma}
\end{equation}
where $TI = \frac{TP}{TP + \alpha \cdot FN + \beta \cdot FP}$ is the Tversky Index with $\alpha=0.01$, $\beta=0.95$, and $\gamma=1.5$. This configuration prioritizes sensitivity over specificity, which is essential for medical segmentation tasks where missing pathological regions is more critical than over-segmentation. The use of task-optimized loss functions enables fair comparison with recent methods that similarly employ specialized losses for imbalanced segmentation.

\begin{table}[t]
\caption{Ablation study results on PROMISE12 dataset. CSP refers to CSP blocks, while dynamic K-NN refers to proposed dynamic K-NN attention mechanism.} 
\footnotesize  
\label{tab:ablation}
\begin{center}       
\begin{tabular}{|c|c|c|c|c|}
\hline
\rule[-1ex]{0pt}{3.5ex} \textbf{Variant} & \textbf {DSC} & \textbf {IoU} & \textbf {HD95} & \textbf {Parameters} \\
\hline
\rule[-1ex]{0pt}{3.5ex} \textbf{Baseline} & 0.8149 & 0.7736 &  8.0724 & 7,849,058 \\
\hline
\rule[-1ex]{0pt}{3.5ex} \textbf{CSP + baseline} & 0.8133 & 0.7684 & 8.0151 & \textbf{6,287,522} \\
\hline
\rule[-1ex]{0pt}{3.5ex} \textbf{Dynamic K-NN + baseline} & 0.8414 & 0.7990 & 6.8908 & 9,242,852 \\
\hline
\rule[-1ex]{0pt}{3.5ex} \textbf{CSP + dynamic K-NN + baseline (KLO-Net)} & \textbf{0.8555} & \textbf{0.8137} & \textbf{6.4355} & 7,681,316 \\
\hline
\end{tabular}
\end{center}
\end{table}

\section{Comparative Analysis Setup}

\subsection{Datasets}

Two public MRI datasets are used for experiments: PROMISE12\cite{Litjens14} and PROSTATEx\cite{Litjens17}. The PROMISE12 dataset is used for ablation studies to analyze the contribution of each component in KLO-Net. The PROSTATEx dataset is used to compare KLO-Net against other state-of-the-art baseline models.

\subsubsection{PROMISE12}
PROMISE12 dataset\cite{Litjens14} contains 80 T2-weighted MRI volumes from multiple centers with different acquisition protocols, making it a challenging benchmark for prostate segmentation. The dataset is split into 50 volumes for training and 30 for testing. In this work, we convert them to 2D axial PNG images for the ablation study.

\subsubsection{PROSTATEx dataset}
The PROSTATEx collection\cite{Litjens17} comprises 204 multi-parametric prostate MRI studies (T2-w, DWI/ADC, DCE, PD) and was originally released for the 2017 SPIE-AAPM-NCI challenge; however, the public release includes only lesion centroids and thumbnail screenshots, with no voxel-level gland annotations, making it unsuitable for direct segmentation training. To obtain ground-truth masks, an open-source PROSTATEx extension by Cuocolo et al.\cite{Cuocolo2021ProstateX, Cuocolo2021DLWZ} is obtained, which provides manually quality-controlled whole-gland and optional zonal masks for all 204 T2-weighted volumes. In this work, the whole-gland masks are used only. The dataset is converted to axial PNGs and split at the per patient level, which into 151 training, 19 validation, and 19 test volumes (3,081 / 385 / 387 axial slices, respectively).

\subsection{Evaluation Metrics}

In order to evaluate model performance, both segmentation accuracy and computational efficiency metrics are used. For the evaluation of the segmentation accuracy, Dice Similarity Coefficient\cite{dice1945, zou2004statistical} (DSC), Intersection over Union\cite{everingham2010pascal} (IoU), and the 95\% Hausdorff Distance\cite{taha2015metrics} (HD95) are used. All evaluations are conducted under 2D predict results.

\subsubsection{Segmentation Accuracy Metrics}
For segmentation accuracy evaluation, DSC, IoU and the HD95 are used. DSC measures the overlap between predicted and ground truth segmentations:
\begin{equation}
\text{DSC} = \frac{2|P \cap G|}{|P| + |G|}
\end{equation}
where $P$ and $G$ represent predicted and ground truth masks respectively. 

IoU, also known as Jaccard index, is calculated as:
\begin{equation}
\text{IoU} = \frac{|P \cap G|}{|P \cup G|}
\end{equation}

The 95\% Hausdorff Distance (HD95) is used to evaluate boundary accuracy, measuring the 95th percentile of surface distances between predicted and ground truth masks.

\subsubsection{Computational Efficiency Metrics}

To evaluate computational efficiency, we measure floating-point operations, i.e., GFLOPs, peak GPU memory usage during inference, total trainable parameter count, and model size in megabytes, i.e., MB, for storage comparison. All experiments, model training, and evaluation are conducted in the same environment setup with a NVIDIA RTX 4090 GPU.

\subsection{Ablation Study}

In order to evaluate the individual contribution of each component in KLO-Net, the ablation study is conducted on the PROMISE12 dataset. Table~\ref{tab:ablation} presents the evaluation results for 4 different configurations. The baseline model is a U-Net variant with encoder starts with 32 filters. Details of different configurations are: (1) baseline model, (2) baseline model with only CSP blocks, (3) baseline model with only dynamic K-NN attention, and (4) the complete KLO-Net architecture.

\begin{figure}[t]
    \centering
    \includegraphics[width=0.82\linewidth]{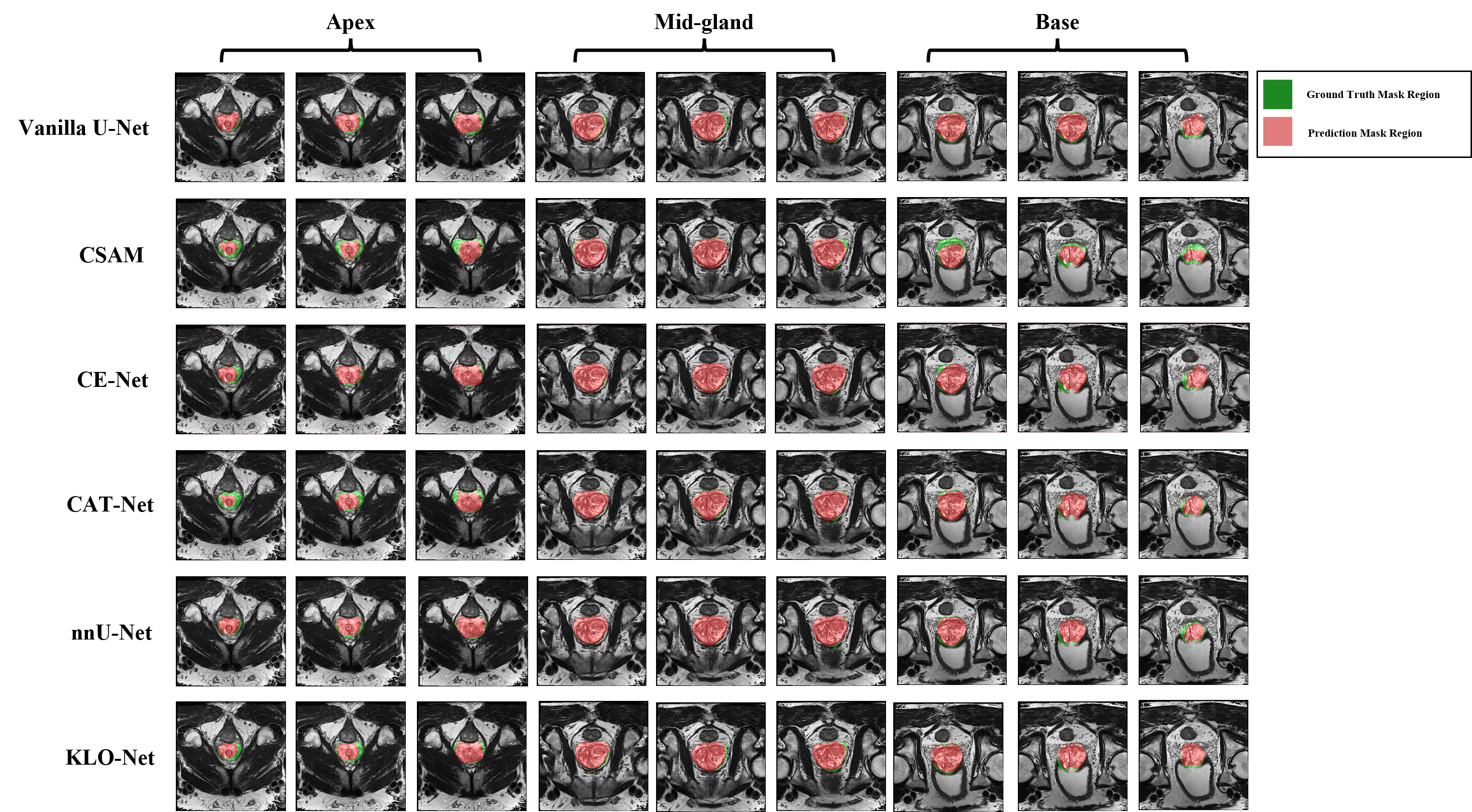}
    \caption{Qualitative comparison of whole-prostate segmentation on a single patient in PROSTATEx dataset}
    \label{fig:Case25-Visulization}
\end{figure}

\subsection{Performance Evaluation on PROSTATEx Dataset}

KLO-Net is compared against five established segmentation methods, i.e., Vanilla U-Net\cite{Ronneberger15}, CSAM\cite{Hung24}, CE-Net\cite{Gu19}, CAT-Net\cite{CAT_Hung23}, and nnU-Net\cite{Isensee21} (2D), on the PROSTATEx dataset to demonstrate its effectiveness in striking a balance between segmentation accuracy and computational efficiency. Table~\ref{tab:comparison} presents comprehensive performance results, including DSC and IoU for segmentation accuracy evaluation, GFLOPs for computational comparison, parameters, and model size to analyze model complexity. Table~\ref{tab:regional_evaluation} presents the results of the regional accuracy evaluation, and Figure~\ref{fig:Case25-Visulization} presents a single patient's qualitative results from the PROSTATEx dataset.

\begin{table*}[t]
\caption{Performance comparison of different models on PROSTATEx dataset. Best results are shown in bold.}
\footnotesize  
\label{tab:comparison}
\begin{center}
\renewcommand{\arraystretch}{1.2}
\begin{tabular}{|l|c|c|c|c|c|c|}
\hline
\textbf{Model} & \textbf{Dice} & \textbf{IoU} & \textbf{GFLOPs} & \textbf{\begin{tabular}[c]{@{}c@{}}GPU Memory\\(Peak/MB)\end{tabular}} & \textbf{Parameters} & \textbf{\begin{tabular}[c]{@{}c@{}}Model Size\\(MB)\end{tabular}} \\
\hline
Vanilla U-Net\cite{Ronneberger15} & 0.8302 & 0.7783 & 40.111 & 198.1 & 17,261,825 & 65.85 \\
\hline
CSAM\cite{Hung24} & 0.8078 & 0.743 & 406.122 & 1142.4 & 139,069,648 & 530.51 \\
\hline
CE-Net\cite{Gu19} & 0.6493 & 0.5881 & 54.52 & 608.59 & 29,003,093 & 110 \\
\hline
CAT-Net\cite{CAT_Hung23} & 0.8189 & 0.7589 & 525.567 & 3022.46 & 614,093,698 & 2342.58 \\
\hline
nnU-Net\cite{Isensee21} (2D) & 0.8421 & 0.7939 & 180.95 & 264.31 & 92,463,532 & 255.53 \\
\hline
\textbf{KLO-Net} & \textbf{0.8567} & \textbf{0.8067}  & \textbf{13.426} &\textbf{186.0} & \textbf{7,681,316} &\textbf{29.30} \\
\hline
\end{tabular}
\end{center}
\end{table*}

\begin{table*}[t]
\caption{Regional Performance Comparison on Apex, Mid-Gland, and Base Regions of PROSTATEx Dataset}
\footnotesize  
\begin{center}
\renewcommand{\arraystretch}{1.65}
\begin{minipage}{0.31\textwidth}
\centering
\textbf{Apex Region}\\
\vspace{0.1cm}
\begin{tabular}{|l|c|c|}
\hline
\textbf{Model} & \textbf{Dice} & \textbf{IoU} \\
\hline
Vanilla U-Net & 0.8475 & 0.8021 \\
\hline
CSAM & 0.8039 & 0.7492 \\
\hline
CE-Net & 0.5696 & 0.5138 \\
\hline
CAT-Net & 0.8026 & 0.7519 \\
\hline
nnU-Net (2D) & 0.8347 & 0.7958 \\
\hline
\textbf{KLO-Net} &\textbf{0.8535}  & \textbf{0.8113} \\
\hline
\end{tabular}
\end{minipage}
\hspace{0.015\textwidth}
\begin{minipage}{0.31\textwidth}
\centering
\textbf{Mid-Gland Region}\\
\vspace{0.1cm}
\begin{tabular}{|l|c|c|}
\hline
\textbf{Model} & \textbf{Dice} & \textbf{IoU} \\
\hline
Vanilla U-Net & 0.9269 & 0.8719 \\
\hline
CSAM & 0.9029 & 0.8335 \\
\hline
CE-Net & 0.9113 & 0.8441 \\
\hline
CAT-Net & 0.9196 & 0.8592 \\
\hline
nnU-Net (2D) & 0.9306 & 0.8772 \\
\hline
\textbf{KLO-Net} & \textbf{0.9323} & \textbf{0.8797}\\ 
\hline
\end{tabular}
\end{minipage}
\hspace{0.015\textwidth}
\begin{minipage}{0.31\textwidth}
\centering
\textbf{Base Region}\\
\vspace{0.1cm}
\begin{tabular}{|l|c|c|}
\hline
\textbf{Model} & \textbf{Dice} & \textbf{IoU} \\
\hline
Vanilla U-Net & 0.6994 & 0.6434 \\
\hline
CSAM & 0.7088 & 0.6373 \\
\hline
CE-Net & 0.4419 & 0.3812 \\
\hline
CAT-Net & 0.7283 & 0.6578 \\
\hline
nnU-Net (2D) & 0.7493 & 0.6964 \\
\hline
\textbf{KLO-Net} & \textbf{0.7738} & \textbf{0.7177} \\ 
\hline
\end{tabular}
\end{minipage}
\end{center}
\vspace{0.2cm}
\label{tab:regional_evaluation}
\end{table*}

\section{Results}

\subsection{Ablation Study}
From Table~\ref{tab:ablation}, i.e., ablation study results, the KLO-Net achieves the highest performance, i.e., 0.8555 DSC and 0.8137 IoU. Compared to the baseline model, the integration of CSP blocks reduced the model's parameter count from 7.85 million to 6.29 million, resulting in a 20\% reduction in model complexity. The complete KLO-Net architecture, combining both CSP and dynamic K-NN attention, results reduced of model complexity with slightly higher DSC and IoU compare to baseline model with same dynamic K-NN attention setup, maintaining a more balanced parameter count, 7.68 million. For HD95, the complete KLO-Net achieved the best HD95 score of 6.4355, which is lower than other configurations.

\subsection{Performance Evaluation}
From Table~\ref{tab:comparison} regarding to performance evaluation, KLO-Net achieves the overall highest DSC of 0.8567 and IoU of 0.8067, outperforming all other models. By comparing efficiency metrics, KLO-Net requires only 13.426 GFLOPs, compared to vanilla U-Net’s 40.111 GFLOPs, representing an approximately 67 percent reduction in computational operations. More complex models, such as CSAM and CAT-Net, demand considerably higher computational resources, at 406.122 and 525.567 GFLOPs, respectively. In terms of model complexity, KLO-Net contains 7.68 million parameters. This is substantially fewer than the vanilla U-Net, which has 17.26 million parameters and occupies 65.85 MB of memory, resulting in a 56\% reduction in model size. The efficiency is also reflected in memory utilization, with KLO-Net requiring only 186.0 MB of peak GPU memory, which is the lowest among all compared methods.

\subsection{Regional Analysis}
From Table~\ref{tab:regional_evaluation}, regarding regional analysis results, KLO-Net consistently achieves the best performance across all three regions: apex region with Dice score of 0.8535, mid-gland region with Dice score 0.9323 , and base region with Dice score of 0.7738. All models achieve their best performance at mid-gland region segmentation and perform poorly on the base region, indicating that models face challenges when dealing with base region segmentation.

\section{Discussion}

The experimental results demonstrate that the proposed KLO-Net effectively balances segmentation accuracy and computational efficiency for prostate gland segmentation from MRI. From the ablation study on the PROMISE12 dataset in Table~\ref{tab:ablation}, both CSP blocks and the dynamic K-NN attention contribute in complementary ways. Introducing CSP into the baseline U-Net encoder substantially reduces the parameter count (from 7.85M to 6.29M) while maintaining very similar DSC and IoU performance. This indicates that CSP successfully removes redundant computation without degrading segmentation quality. In contrast, adding the dynamic K-NN attention to the baseline improves DSC and IoU and reduces HD95 at the cost of increased parameters. When both components are combined in KLO-Net, the model achieves the highest accuracy across all metrics while keeping a parameter count close to the baseline. This confirms that CSP and dynamic K-NN attention act synergistically rather than redundantly.

The comparative evaluation on the PROSTATEx dataset highlights the trade-off between accuracy and efficiency. As shown in Table~\ref{tab:comparison}, KLO-Net achieves the highest Dice and IoU among all compared methods while requiring the lowest GFLOPs and the smallest peak GPU memory usage. In particular, KLO-Net achieves better segmentation performance than 2D nnU-Net and vanilla U-Net with substantially fewer parameters and a much smaller model size. These results indicate that inserting sparse, content-adaptive attention only at deeper encoder stages and the bottleneck is sufficient to capture the long-range dependencies that are critical for prostate delineation, without incurring the high computational overhead associated with full self-attention or heavy transformer-based architectures. At the same time, the CSP encoder reduces redundancy in convolutional blocks, making KLO-Net more suitable for deployment on resource-constrained clinical workstations.

The regional analysis on the PROSTATEx dataset provides additional insight. Table~\ref{tab:regional_evaluation} shows that KLO-Net consistently achieves the best Dice and IoU across the apex, mid-gland, and base regions. All models achieve their best performance in the mid-gland region and perform worse in the base region, which reflects known anatomical variability and ambiguous boundaries at the base. However, KLO-Net shows the largest relative gains in the base region, suggesting that dynamically adjusting the attention density is helpful in anatomically complex or low-contrast areas. By allowing each spatial location to adapt its number of attention connections, the dynamic K-NN module can allocate denser attention to boundary regions and sparser attention to homogeneous background, leading to more robust boundary delineation. Qualitative examples in Fig.~\ref{fig:Case25-Visulization} support this observation, where KLO-Net better follows the gland contour and reduces under- and over-segmentation compared to other methods.

\section{Conclusion}

In this study, we propose KLO-Net, a novel U-Net architecture integrating dynamic K-NN attention and CSP blocks for efficient prostate gland segmentation from MRI. Compared to five state-of-the-art methods on the PROSTATEx dataset, KLO-Net substantially reduces computational demands and model complexity while achieving higher segmentation accuracy. The adaptive attention mechanism of KLO-Net proves effective in accurately segmenting complex prostate regions such as the apex and base, as confirmed by the regional analysis. In addition, its lightweight design and low memory footprint indicate strong potential for integration on edge devices and clinical workstations with limited resources. Future research directions include extending the architecture to 3D or 2.5D segmentation, incorporating multi-modal MRI inputs, and evaluating its generalizability across diverse medical imaging datasets and additional anatomical targets.

\section*{Data availability}
This study used publicly available datasets including PROMISE12 and PROSTATEx.
No new patient data were collected.

\section*{Ethics statement}
All experiments were conducted on publicly available, de-identified datasets.
No human subjects were recruited and no additional IRB approval was required for this study.

\section*{Disclaimer}
This work is for research purposes only and is not intended for clinical diagnosis or medical decision-making.

\bibliographystyle{spiebib}
\bibliography{report}

\end{document}